\documentclass[runningheads]{llncs}
\usepackage[T1]{fontenc}
\usepackage{graphicx}
\begin{document}
\title{Educational-Psychological Dialogue Robot Based on Multi-Agent Collaboration}
%
%
\author{Shiwen Ni \and
Min Yang\inst{*}}

\institute{Shenzhen Institute of Advanced Technology, Chinese Academy of Sciences \\
\email{\{sw.ni,min.yang\}@siat.ac.cn}}
\maketitle              
\begin{abstract}
Intelligent dialogue systems are increasingly used in modern education and psychological counseling fields, but most existing systems are limited to a single domain, cannot deal with both educational and psychological issues, and often lack accuracy and professionalism when dealing with complex issues.
To address these problems, this paper proposes an intelligent dialog system that combines educational and psychological counseling functions. The system consists of multiple AI agent, including security detection agent, intent identification agent, educational LLM agent, and psychological LLM agent, which work in concert to ensure the provision of accurate educational knowledge Q\&A and psychological support services. Specifically, the system recognizes user-input intentions through an intention classification model and invokes a retrieval-enhanced educational grand model and a psychological grand model fine-tuned with psychological data in order to provide professional educational advice and psychological support.

\end{abstract}
\section{Introduction}
In the field of modern education and psychological counseling, the application of intelligent dialog systems is becoming more and more widespread. Educational intelligent dialog systems can assist students in answering questions, providing learning support and improving learning efficiency. Psychological counseling intelligent dialog systems, on the other hand, can provide users with emotional support and help them relieve psychological pressure \cite{zhang2024cpsycoun}. However, most of the existing intelligent dialog systems are limited to a single domain and cannot handle both educational and psychological issues \cite{dan2023educhat,dinh2023educhat,oster2024chatgpt}. In addition, these systems often lack accuracy and professionalism when dealing with complex issues.

Currently, large-scale pre-trained model \cite{vaswani2017attention,chu2024history} techniques have made significant progress in semantic understanding and text generation, and are widely used in various natural language processing tasks. These models can effectively understand the context, but still suffer from insufficient answer accuracy and expertise when used alone. To address these problems, retrieval enhancement techniques have been introduced to significantly improve the quality and richness of the model's answers by incorporating external knowledge bases (e.g., Baidu Encyclopedia, Wikipedia). However, these techniques are still insufficiently applied in existing intelligent dialog systems, especially in the combined application in the fields of education and psychological counseling, for which there is no mature solution yet.

In the field of education, students not only need answers to subject knowledge in the learning process, but also may face psychological pressure and emotional distress, in which case a single educational dialog system cannot provide comprehensive help. Similarly, in the field of psychological counseling, users may also need guidance and advice on education while seeking emotional support, and existing psychological dialog systems are not capable of handling such issues. Based on the above background, we propose an intelligent dialog bot that combines educational and psychological counseling functions. It recognizes user-inputted intent through an intent classification model, and invokes an educational grand model based on retrieval enhancement and a psychological grand model fine-tuned with psychological data, respectively, so as to provide professional educational counseling and psychological support. The proposed intelligent dialog bot not only solves the deficiencies in the prior art, but also improves the user experience and provides more comprehensive and accurate services.

\section{Methodology}
In this paper, we propose an intelligent dialogue system that combines educational and psychological LLMs, and the system consists of multiple AI agents together. As shown in Fig. 1, the user's input is first filtered by a security detection agent, and then the intent of the user input is identified by an intent identification agent, and the different intents are fed to the corresponding LLM agents for processing, respectively, and the system is inter-systemmed through the four different agents in order to provide accurate educational knowledge Q\&A and psychological support services.
\begin{figure}
\includegraphics[width=\textwidth]{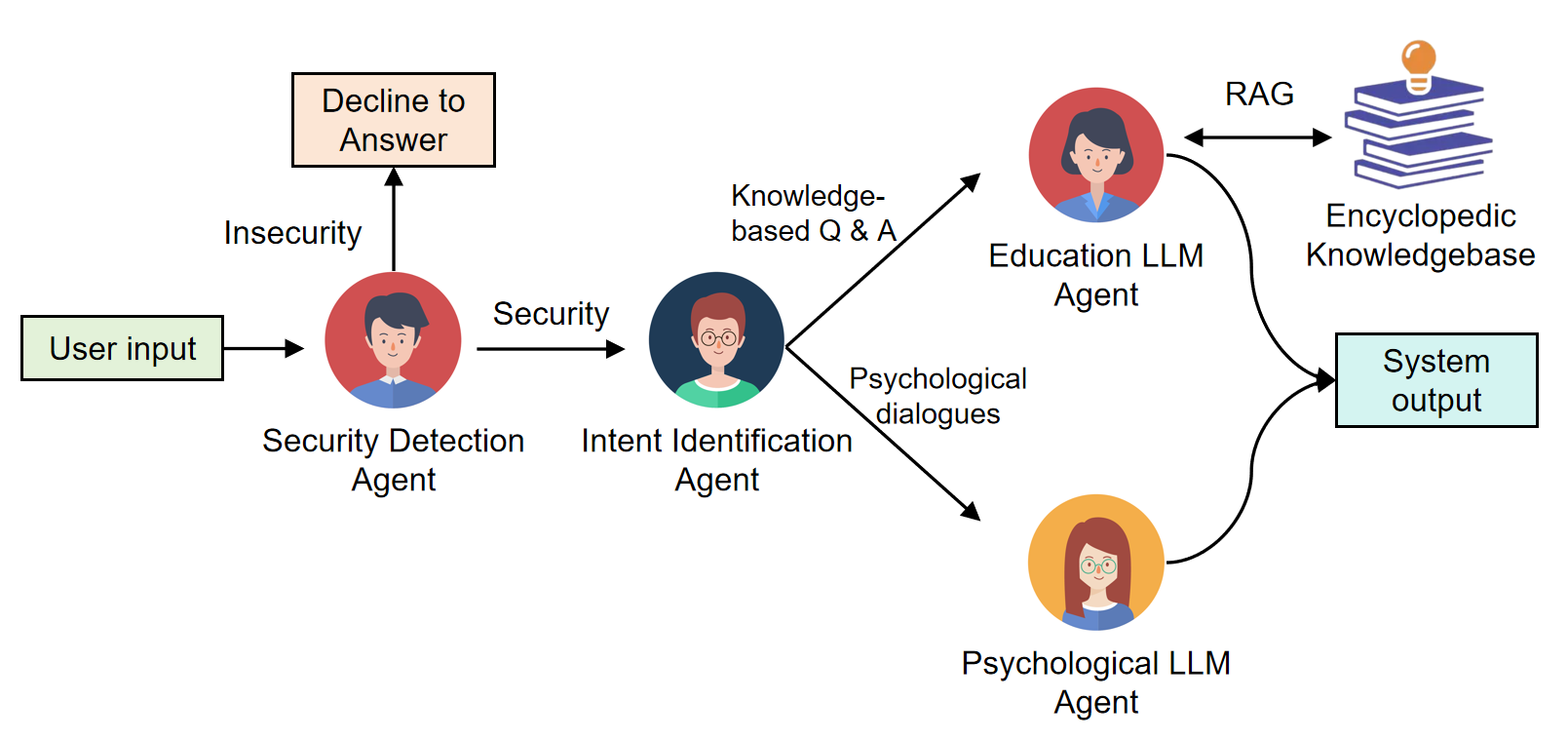}
\caption{A figure caption is always placed below the illustration.
Please note that short captions are centered, while long ones are
justified by the macro package automatically.} \label{fig1}
\end{figure}

\textbf{Security Detection Agent}: We use a safety model to determine the safety of input content and filter out unsafe inputs such as insults discrimination, illegal crimes, political sensitivity, violent tendencies, etc. to ensure that the model does not respond to unsafe questions. The safety model is a binary classification model that uses bert-base-chinese \cite{devlin2018bert} as a base model and then fine-tuned using 40k normal questions as positive examples and 40k risky questions as negative examples to ultimately implement a safety classification model. The model uses bert-base-chinese as a base model, which ensures a reliable response time. It also uses 40k positive and negative examples as a dataset to ensure the accuracy of the model. The security model is in the first loop in this system, and all inputs are judged by the security model first, and only those that meet the requirements are allowed to go to the later links.

\textbf{Intent Identification Agent}: this system designs a BERT-based \cite{devlin2018bert} binary classification model for accurately recognizing the input content of student users and determining whether it should be processed by the psychological or educational LLM. To achieve this goal, a large amount of data including 20,000 educational questions and 700,000 psychological questions were collected as positive and negative examples, respectively. Such a large dataset ensures that the model can cover a wide range of problem types and diverse user needs. 

To address the sample imbalance problem, we adopt Focal Loss as the loss function of the classification model, which effectively reduces the impact of simple samples on the total loss by introducing a dynamically adjusted weighting mechanism. This mechanism allows the model to focus on those difficult samples during training, which significantly improves the classification performance in the case of category imbalance.This feature of Focal Loss ensures that the model maintains high accuracy when dealing with the problem of a small number of education classes.
In terms of model selection, we choose bert-base-chinese as the base model. Based on the bert-base-chinese model, we added a linear layer to it for further training, and finally constructed an efficient educational-psychological binary classification model.

\textbf{Educational LLM Agent}: we developed a retrieval-enhanced \cite{huang2024survey} educational large language model. First, the entries in Baidu Encyclopedia are converted into vectors in advance using bge-large-zh \cite{xiao2023c} as the embedding model, and these vectors are imported into the faiss vector database, for which an efficient HNSW (Hierarchical Navigable Small World) indexing structure is built. This process ensures efficient storage and fast retrieval of data.
When a user enters a question, the system uses the embedding model to transform the question into a vector representation, and then searches the educational vector database to quickly find the 100 most relevant data to the user's question. These initially filtered data are able to cover multiple aspects of the user's question, providing a wide range of contextual information. To further improve the accuracy and relevance of the answers, we use bge-reranker-large as a rerank model to reorder these 100 pieces of data at a finer granularity. the rerank model selects the 3 most useful pieces of data by deeply analyzing the content of each piece of data and how well it matches the user's question. These data will be used as the optimal context of the question and added to the prompt (prompt), which greatly improves the quality and accuracy of the answers generated by the large model.

 In addition, our educational large language model is based on the fine-tuning of the Qwen1.5-7B model \cite{bai2023qwen} on the COIG-CQIA \cite{bai2024coig} dataset, which is specifically optimized to fit the needs of the Chinese education domain.
 In summary, by transforming the massive data of Baidu Encyclopedia into an educational vector database and combining HNSW indexing, embedding model, rerank model, and fine-tuned Qwen1.5-7B LLM, we constructed an efficient and accurate retrieval-enhanced educational large model. This model not only responds quickly to user-input questions, but also provides high-quality, detailed and relevant answers, which greatly improves the user experience and learning effect.

\textbf{Psychological LLM Agent}: In order to build a powerful psychological counseling grand model, we manually combined GPT-4 \cite{brown2020language} to generate a large amount of Chinese and English multi-round psychological conversation data. These data cover a wide range of counseling scenarios and issues, enabling the model to provide effective psychological support in multiple contexts. We used these generated data to fine-tune the base model to enhance its performance in psychological counseling.

The base model we chose is Qwen1.5-7B-chat its strong language comprehension and generation capabilities, and it performs particularly well in dialog systems. Through the fine-tuning technique, we further optimize the model's ability to handle counseling conversations while retaining its original advantages. Fine-tuning is able to improve the model's performance in specific domains without significantly increasing the computational overhead, making it more suitable for dealing with counseling-type problems.
As a result of this fine-tuning process, the model was significantly improved in terms of its counseling capabilities. Specifically, the model is able to understand the user's psychological needs more accurately, provide more attentive and effective suggestions, and maintain coherence and consistency across multiple rounds of dialog. This enables the model to better support users in practical applications and provide high-quality psychological counseling services.
\begin{table}[]
\centering
\begin{tabular}{|l|c|c|c|c|c|}
\hline
Model                       & Chinese & Mathematics & English & Science & Ethics \\ \hline
ChatGLM3-6B                 & 56.2    & 39.2        & 62.7    & 71.9    & 83.9   \\ \hline
Qwen1.5-7B            & 73.9    & 72.5        & 80.8    & 78.0    & 93.1   \\ \hline
GPT-4                        & 70.8    & 69.6        & 92.5    & 85.3    & 94.2   \\ \hline
Educational LLM Agent (our) & 75.3    & 73.2        & 80.9    & 80.4    & 94.9   \\ \hline
\end{tabular}
\caption{Experimental results on E-Eval primary school subjects.}
\label{t1}
\end{table}
\section{Experiment}
We validated the effectiveness of our educational LLM agent using the E-EVAL \cite{hou2024eval} benchmark. E-EVAL is the first comprehensive assessment benchmark customized for K-12 education in China. E-EVAL comprises 4,351 multiple-choice questions spanning primary, middle, and high school levels, covering a diverse array of subjects. In our comparison, we evaluated the performance of ChatGLM3-6B \cite{glm2024chatglm}, Qwen1.5-7B \cite{bai2023qwen}, GPT-4 \cite{brown2020language}, and our own Educational LLM Agent. Initially, we present the results from experiments on primary subjects in Table 1. The results indicate that our Educational LLM agent performs exceptionally well, surpassing both Qwen1.5-7B and ChatGLM3-6B. Notably, in subjects such as Chinese and Ethics, our agent even outperformed GPT-4.

\begin{table}[t]
\centering
\resizebox{350pt}{!}{
\begin{tabular}{|l|c|c|c|c|c|c|c|c|c|}
\hline
Model                       & Chinese & Math & English & \multicolumn{1}{l|}{Physics} & \multicolumn{1}{l|}{Chemistry} & Biology & Politics & History & Geography \\ \hline
ChatGLM3-6B                 & 52.0    & 37.2        & 70.2    & 69.7                         & 67.5                           & 72.1    & 82.7     & 81.1    & 70.1      \\ \hline
Qwen1.5-7B             & 68.2    & 68.6        & 89.1    & 68.4                         & 80.4                           & 80.4    & 90.2     & 89.7    & 90.0      \\ \hline
GPT4                        & 54.7    & 59.8        & 93.4    & 76.3                         & 67.5                           & 87.5    & 83.9     & 88.9    & 81.9      \\ \hline
Our & 70.4    & 68.9        & 89.3    & 69.4                         & 81.2                           & 83.1    & 92.2     & 94.5    & 93.7      \\ \hline
\end{tabular}
}
\caption{Experimental results on E-Eval middle school subjects.}
\label{t2}
\end{table}

The results of the experiment for middle school subjects are shown in Table 2. The experimental results show that our Educational LLM agent outperforms the GPT-4 in the English subject and performs optimally in all other subjects. In language subjects our intelligences achieved an accuracy rate of 70.4\%, far exceeding the 54.7\% of the GPT-4. Improvements were evident in politics, history, and geography, but less so in math, physics, and chemistry, indicating that the RAG was more effective in improving liberal arts skills.

The results of the experiment for high school subjects are shown in Table 3. The overall performance of LLMs in high school subjects was somewhat lower than in middle school because high school is more difficult. Similarly except for English subject where our educational proxy is below GPT-4, we have the best performance in all subjects. Our method continues to perform very brightly in language and literature, for example, with an accuracy rate of 67.5\% in language and a staggering 83.5\% in history.

It should be noted that we are a whole multi-agent collaborative system, not a single LLM, so our system can have better security, as well as the ability to call different agents to meet the educational knowledge quiz and professional counseling.
\begin{table}[]
\centering
\resizebox{350pt}{!}{
\begin{tabular}{|l|c|c|c|c|c|c|c|c|c|}
\hline
Model       & Chinese & Mathematics & English & \multicolumn{1}{l|}{Physics} & \multicolumn{1}{l|}{Chemistry} & Biology & Politics & History & Geography \\ \hline
ChatGLM3-6B & 40.5    & 33.8        & 64.7    & 52.1                         & 51.1                           & 56.1    & 70.1     & 68.5    & 57.0      \\ \hline
Qwen1.5-7B  & 62.9    & 43.8        & 80.0    & 58.9                         & 71.0                           & 73.0    & 80.6     & 79.7    & 77.6      \\ \hline
GPT4        & 39.3    & 42.6        & 88.5    & 61.5                         & 59.0                           & 63.8    & 65.5     & 78.2    & 78.8      \\ \hline
Our         & 67.5    & 44.8        & 81.1    & 62.2                          & 72.1                           & 77.4    & 82.3     & 83.5    & 80.7      \\ \hline
\end{tabular}}
\caption{Experimental results on E-Eval high school subjects.}
\label{t3}
\end{table}
\section{Conclusion}
In this study, we developed an educational-counseling dialog robot based on multi-agent collaboration, which breaks through the limitations of existing single-domain intelligent dialog systems by combining educational and counseling functions. Our experimental results on the E-EVAL benchmark test show that our educational LLM agent outperforms existing models in several subjects, including Qwen1.5-7B and ChatGLM3-6B, and even exceeds GPT-4 in subjects such as Chinese and ethics.
Our system demonstrated a high degree of safety, accuracy, and professionalism, validating the effectiveness of multi-intelligence collaboration in providing comprehensive educational and counseling services. 

\section*{Acknowledgement}
This work was supported by China Postdoctoral Science Foundation (2024M753398), Postdoctoral Fellowship Program of CPSF (GZC20232873), GuangDong Basic and Applied Basic Research Foundation (2023A1515110718 and 2024A1515012003).
%
%
%
\bibliography{mybibliography}

\end{document}